\documentclass{article} 
\usepackage{iclr2026_conference,times}

\usepackage{amsmath,amsfonts,bm}

\def\eqref#1{equation~\ref{#1}}

\def\1{\bm{1}}

\DeclareMathAlphabet{\mathsfit}{\encodingdefault}{\sfdefault}{m}{sl}
\SetMathAlphabet{\mathsfit}{bold}{\encodingdefault}{\sfdefault}{bx}{n}

\usepackage{hyperref}
\usepackage{url}
\usepackage{multirow}
\usepackage{booktabs,wrapfig}
\usepackage{graphicx}
\usepackage{subcaption}
\usepackage{algorithm}
\usepackage{listings}
\usepackage{algpseudocode}
\usepackage{adjustbox}
\usepackage{enumitem}

\usepackage{marvosym}

\usepackage{caption}
\usepackage{colortbl}

\usepackage{soul}

\newcommand{\hc}{\cellcolor{teal!20}}

\hypersetup{
	colorlinks=true,
	citecolor=[rgb]{0,0.08,0.45},
}
\makeatletter
\def\thanks#1{\protected@xdef\@thanks{\@thanks
        \protect\footnotetext{#1}}}
\makeatother

\usepackage{amsmath}
\usepackage{bm}
\usepackage{graphicx}
\usepackage{multirow}
\usepackage{pifont}
\let\oldding\ding
\renewcommand{\ding}[2][1]{\scalebox{#1}{\oldding{#2}}}

\usepackage{amsthm}
\newtheorem{assumption}{Assumption}
\newtheorem{lemma}{Lemma}

\usepackage{subcaption}
\usepackage{booktabs}
\usepackage[table]{xcolor}
\usepackage{wrapfig}
\usepackage{enumitem}
\usepackage{soul}

\usepackage{makecell}

\usepackage{amsthm}
\newtheorem{remark}{Remark}

\usepackage{amssymb}
\usepackage{amsmath}
\makeatletter
\def\thanks#1{\protected@xdef\@thanks{\@thanks
        \protect\footnotetext{#1}}}
\makeatother

\usepackage{algorithm}
\usepackage{hyperref}
\usepackage{url}
\usepackage{multirow}
\usepackage{booktabs}
\usepackage{graphicx}
\usepackage{subcaption}
\usepackage{algorithm}
\usepackage{listings}
\usepackage{algpseudocode}
\usepackage{adjustbox}
\usepackage{enumitem}

\title{QFT: Quantized Full-parameter Tuning of LLMs with Affordable Resources}

\author{Zhikai Li\textsuperscript{1}\ , Xiaoxuan Liu\textsuperscript{2}\ , Banghua Zhu\textsuperscript{2}\ , Zhen Dong\textsuperscript{2\Letter}, Qingyi Gu\textsuperscript{1\Letter}\thanks{\textsuperscript{\Letter} Corresponding authors.}, Kurt Keutzer\textsuperscript{2}\\
\textsuperscript{1}Institute of Automation, Chinese Academy of Sciences \quad
\textsuperscript{2}University of California, Berkeley\\
\footnotesize{\texttt{\{zhikai.li,qingyi.gu\}@ia.ac.cn}}\\
\footnotesize{\texttt{\{xiaoxuan\_liu,banghua,zhendong,keutzer\}@berkeley.edu}}\\
}

\iclrfinalcopy 
\begin{document}

\maketitle

\begin{abstract}
Large Language Models (LLMs) have showcased remarkable impacts across a wide spectrum of natural language processing tasks.  
Fine-tuning these pretrained models on downstream datasets provides further significant performance gains; however, this process typically requires a large number of expensive, high-end GPUs. Although there have been efforts focused on parameter-efficient fine-tuning, they cannot fully unlock the powerful potential of full-parameter fine-tuning.
In this paper, we propose QFT, a Quantized Full-parameter Tuning framework for LLMs that quantizes and stores all training states, including weights, gradients, and optimizer states, in INT8 format to reduce training memory, thereby enabling full-parameter fine-tuning on existing GPUs at an affordable cost.
To ensure training performance, we make two key efforts: i) for quantized gradients and optimizer states, we theoretically prove that the Lion optimizer, with its property of consistent update magnitudes, is highly robust to quantization; ii) and for quantized weights, we employ the hybrid feature quantizer, which
 identifies and protects a small subset of sparse critical features while quantizing the remaining dense features, thus ensuring accurate weight updates without FP32 backups.
Moreover, to support backpropagation in the integer context, we develop a stack-based gradient flow scheme with O(1) complexity, forming a unified integer training pipeline.
As a result, QFT reduces the model state memory to 21\% of the standard solution while achieving comparable performance, e.g., tuning a LLaMA-7B model requires only $<$30GB of memory, making it feasible on a single A6000 GPU.
\end{abstract}

\section{Introduction}
Large Language Models (LLMs), with up to hundreds of billions of parameters, have left an indelible mark on the landscape of natural language processing tasks, showcasing their remarkable impacts across a diverse spectrum of applications and domains~\citep{touvron2023llama1,touvron2023llama,brown2020language,zhang2022opt}. Fine-tuning these pre-trained models on downstream datasets enhances their ability to understand and perform specific tasks~\citep{zhao2023survey}.
However, due to the enormous number of parameters, the fine-tuning process relies on massive and expensive GPU resources, resulting in extremely high costs.

Parameter-efficient fine-tuning (PEFT), which tunes only a subset of parameters, is considered a practical choice in low-resource settings~\citep{ding2022delta,hu2021lora,li2021prefix,zhang2025train}. However, due to the limited representational capacity of the smaller parameter set, its performance often falls short of expectations~\citep{lv2023full,zhao2024galore}. Consequently, we focus on full-parameter fine-tuning, with an emphasis on exploring memory optimization strategies to make it feasible on cost-effective resources.

\begin{figure}[t]
    \centering
    \includegraphics[width=0.95\linewidth]{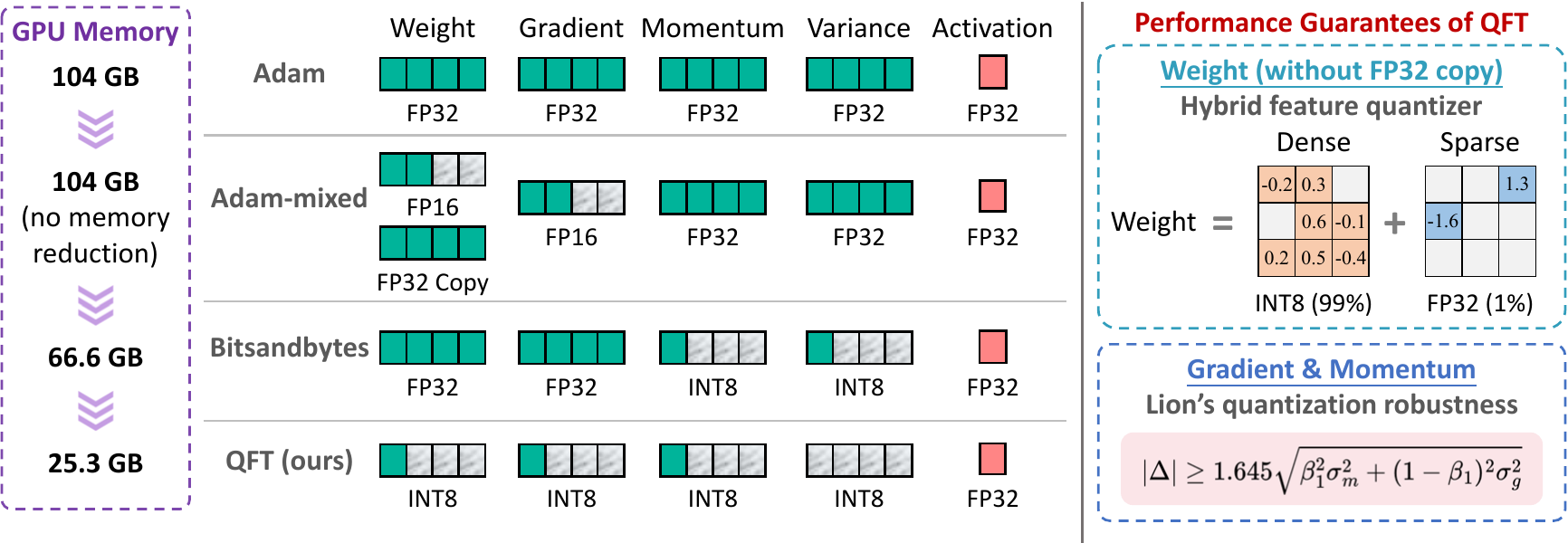}
    \vspace{-0.1cm}
    \caption{Comparison in GPU memory usage of different full-parameter fine-tuning methods. QFT significantly reduces training memory, enabling fine-tuning with affordable resources. To ensure the performance of quantized fine-tuning, QFT adopts the hybrid feature quantizer for weights, and for gradients and momentum, we theoretically prove that Lion exhibits high robustness to quantization, thereby ensuring comparable convergence to FP32 Adam.}
\label{fig:overview}
\vspace{-0.3cm}
\end{figure}

We examine the full spectrum of memory usage in full-parameter fine-tuning, which can be categorized into three components: model states, activation, and other temporary or unusable memory, as shown in Figure \ref{fig:overview}. Model states, which include weights, gradients, and optimizer states (e.g., momentum and variances in Adam~\citep{kingma2015adam}), consume the majority of the memory~\citep{rajbhandari2020zero}.
For instance, when employing the standard FP32 setting with Adam, the memory allocation for weights, gradients, momentum, and variances amounts to 4 times the number of parameters. 
As a result, tuning a LLaMA-7B model necessitates a minimum of 104GB of RAM, which presents a formidable challenge given the limitations of current GPU capacities. 
Notably, although mixed-precision training~\citep{micikevicius2017mixed} reduces computation precision, it requires storing an additional FP32 weight copy, hence it only accelerates training but fails to address the memory consumption issues.

In this paper, we introduce QFT, a Quantized Full-parameter Tuning framework for training memory optimization. 
specifically, QFT quantizes and stores all training states in INT8 format, significantly reducing memory consumption and enabling full-parameter fine-tuning on existing GPUs at an affordable cost.
In contrast to traditional quantization-aware training (QAT), QFT focuses specifically on memory optimization by \emph{storing} all parameters in low-bit format. 
To maintain training performance, we make two key efforts: 
\raisebox{-0.8pt}{\ding[1.1]{182\relax}} \textbf{For quantized gradients and optimizer states}, we theoretically prove that the Lion optimizer~\citep{chen2302symbolic}, which tracks only momentum and produces consistent update magnitudes, exhibits strong robustness to quantization. Thus, we employ the Lion optimizer to minimize the effects of quantization of gradients and momentum.
\raisebox{-0.8pt}{\ding[1.1]{183\relax}}  \textbf{For quantized weights}, we employ the hybrid feature quantizer, which selectively retains a small set of sparse critical features while quantizing the majority of dense features, thereby ensuring accurate weight updates. It does not rely on FP32 backups, thus achieving better memory efficiency than mixed-precision training.
On this basis, to enable integer-based backpropagation, we design a stack-based gradient flow scheme with O(1) complexity, constructing a unified integer training pipeline.
It is also worth noting that QFT adopts the INT8 format by default, which is broadly supported by most hardware. This design avoids reliance on specialized data types like FP8~\citep{micikevicius2022fp8} that require high-end GPUs, thus enabling full utilization of existing mid- and low-end GPUs.
Our contributions can be summarized as follows:
\vspace{-5pt}
\begin{itemize}[left=0pt]
    \item  We propose QFT, a Quantized Full-parameter Tuning framework for LLMs. It achieves training memory optimization by reducing storage precision, enabling full-parameter fine-tuning on affordable resources, which separates it from traditional QAT. 
    In addition, QFT offers strong compatibility and can be seamlessly integrated into mainstream LLM training tools.
    \item  To ensure training performance, we first theoretically prove the robustness of the Lion optimizer to quantization, ensuring the reliability of \emph{quantized gradients and optimizer states}. Then, we protect a small subset of critical features within the weights while quantizing the remaining dense features, effectively preserving the accuracy of \emph{quantized weight updates}. Moreover, we develop a stack-based gradient flow scheme with O(1) complexity to enable integer backpropagation.
    \item  We perform instruction tuning on the pre-trained LLaMA-2 models and extensively evaluate performance on various benchmarks. The results demonstrate that our QFT, with memory usage reduced to 21\%, achieves comparable performance to standard floating-point training.
\end{itemize}

\section{Related Work}

\textbf{Efficient Optimizer }
The Adam family optimizers are widely used for training Transformer models \citep{kingma2015adam,loshchilov2017decoupled}, but their optimizer states (momentum and variance) introduce substantial memory overhead. To mitigate this, several memory-efficient alternatives have been proposed. LOMO \citep{lv2023full} uses vanilla SGD, but suffers from slow convergence and poor stability \citep{li2023memory}. BAdam \citep{luo2024badam} adopts a block coordinate descent framework with Adam-style updates. Adafactor \citep{shazeer2018adafactor} reduces memory by storing only aggregated statistics, but remains prone to instability.
In this work, we adopt the Lion optimizer \citep{chen2302symbolic}, which tracks only momentum while achieving performance comparable to Adam. More importantly, its sign-based update yields consistent update magnitudes, making it particularly suitable for robust quantization of gradients and optimizer states.

\textbf{Quantization for Memory Optimization }
Most existing quantization methods focus on inference efficiency \citep{gholami2022survey,dong2019hawq,dong2020hawq,kim2023squeezellm,li2022patch,li2022repq,li2022vit,jacob2018quantization}. This line of work differs fundamentally from traditional QAT \citep{jacob2018quantization,liu2023llm}, which inserts fake quantization nodes during training while keeping parameter storage and computation in floating point, and thus does not reduce training memory usage.
In contrast, quantization-based memory optimization aims to store parameters in low precision to directly reduce training memory. For example, Bitsandbytes \citep{dettmers20218} applies block-wise quantization to compress optimizer states, and QLoRA \citep{dettmers2023qlora} stores frozen pre-trained weights in quantized form while keeping adapters in floating point. In this work, we focus on full-parameter fine-tuning and quantize all model states to achieve comprehensive memory savings without sacrificing fine-tuning performance.

\textbf{Other Memory Optimization Methods }
Previous work has focused on reducing activation memory, including activation offloading \citep{huang2020swapadvisor, wang2018superneurons, peng2020capuchin} and gradient checkpointing \citep{chen2016training, kumar2019efficient, jain2020checkmate, kirisame2020dynamic}. Activation offloading moves activations to external memory at the cost of data transfer, while gradient checkpointing discards activations in the forward pass and recomputes them during backpropagation.
There are also customized schemes for other training states. GaLore \citep{zhao2024galore} reduces gradient memory via low-rank projection, and LOMO \citep{lv2023full} fuses gradient computation and parameter update, reducing gradient memory to $O(1)$. However, LOMO is incompatible with gradient accumulation for large batch sizes, limiting it to unstable small-batch training. In contrast, our framework is orthogonal to and compatible with all these methods.

\section{Methodology}

\subsection{Lion Optimizer: Robust Quantization of Gradients and Momentum}
\label{method:lion}

In contrast to Bitsandbytes~\citep{dettmers20218} which adopts advanced quantization strategies, we prioritize ease of use by applying the simplest uniform quantizer to gradients and momentum, defined as follows:
\begin{align}
Quant: \bm{X}^{(\mathbb{Z})} = \text{clip}\left(\left\lfloor \frac{\bm{X}}{s} \right\rceil+z, 0, 2^b-1 \right), \quad
  De\text{-}quant: \hat{\bm{X}} = s\left(\bm{X}^{(\mathbb{Z})}-z\right) \approx \bm{X},
\label{eq:quant}
\end{align}
where $\bm{X}$ is the floating-point vector, $\bm{X}^{(\mathbb{Z})}$ is the quantized integer vector, $\left\lfloor\cdot\right\rceil$ denotes the round function, and $b \in \mathbb{N}$ is the quantization bit-width. $s\in \mathbb{R}^+$ and $z \in \mathbb{Z}$ are the quantization scale and zero-point, which are determined by the arithmetic lower and upper bounds of $\bm{X}$ as follows:
\begin{equation}
\label{eq:sz}
s = \frac{\max(\bm{X})-\min(\bm{X})}{2^b-1}, \quad z = \left\lfloor-\frac{\min(\bm{X})}{s} \right\rceil.
\end{equation}

The simplicity of the quantizer effectively ensures computational efficiency but also raises concerns about potential impacts on training performance. Fortunately, Lion~\citep{chen2302symbolic} (detailed in Appendix \ref{app:sec1}), which tracks only momentum and applies updates of consistent magnitude to each parameter via the sign operation, can significantly mitigate the adverse effects of quantization. More specifically, Lion has inherent advantages for quantized fine-tuning as follows:
\begin{itemize}[left=0pt]
\vspace{-5pt}
\item \textbf{Simplicity:} Lion only keeps track of the momentum, which saves memory by avoiding storing variances, while eliminating the potential effect of quantized variances.
\item \textbf{Consistent Update Magnitudes:} Lion ensures that updates have the same magnitude for each parameter, which is determined through the sign operation. In a quantized setting, this consistency can mitigate potential imbalances or inaccuracies introduced by limited precision.
\end{itemize}

For the consistent update magnitude property, we provide a detailed proof below to demonstrate Lion's high robustness to quantization of gradients and momentum.

\begin{assumption} 
\label{assump}
Assume that:

\begin{itemize}[left=0pt]
\vspace{-5pt}
    \item (Additive quantization error) The quantization error is additive, i.e., $\hat{x}=x+\delta_x$.
    \item (Bounded Gaussian error) The quantization error $\delta_x$ follows a Gaussian distribution $\mathcal{N}(0,\sigma_x^2)$, where $\sigma_x^2$ is small and bounded.
\end{itemize}
\end{assumption}

\begin{lemma}
    Under Assumption \ref{assump}, when quantizing gradients and momentum in Lion, if the increment $\Delta$ satisfies $|\Delta|\ge 1.645\sqrt{\beta_1^2\sigma_m^2+(1-\beta_1)^2\sigma_g^2}$, then with 95\% probability, $\mathrm{sign}(\Delta)$ remains invariant under quantization.
\end{lemma}

\begin{proof}
    Given Assumption \ref{assump}, we have:
    \begin{align*}
        \hat{m}=m+\delta_m, \; \hat{g}=g+\delta_g, \quad \text{where} \;\; \delta_m\sim \mathcal{N}(0,\sigma_m^2), \; \delta_g\sim \mathcal{N}(0,\sigma_g^2).
    \end{align*}
    When quantizing gradients and momentum, the increment becomes:
    \begin{align*}
        \Delta' = \beta_1\hat{m}+(1-\beta_1)\hat{g} = \Delta + \beta_1\delta_m+(1-\beta_1)\delta_g.
    \end{align*}
    Let $\delta_\Delta=\beta_1\delta_m+(1-\beta_1)\delta_g$. Since $\delta_m$ and $\delta_g$ are independent Gaussian distributions, based on homogeneity and 	additivity, $\delta_\Delta$ follows a Gaussian distribution:
    \begin{align*}
        \delta_\Delta \sim \mathcal{N}(0,\sigma_\delta^2) := \mathcal{N}(0,\beta_1^2\sigma_m^2+(1-\beta_1)^2\sigma_g^2).
    \end{align*}
    To ensure that the sign of $\Delta$ remains invariant after quantization, we require that $\mathrm{sign}(\hat{\Delta}) = \mathrm{sign}(\Delta)$, i.e., $\Delta$ and $\Delta + \delta_\Delta$ must have the same sign.  
    First, consider the case $\Delta > 0$, we require $\Delta + \delta_\Delta > 0 \Leftrightarrow \delta_\Delta > -\Delta$, thus the probability of sign preservation is:
    \begin{align*}
        P(\delta_\Delta > -\Delta) = 1 - P(\delta_\Delta \leq -\Delta).
    \end{align*}
    Standardizing $\delta_\Delta$ yields:
    \begin{align*}
        P(\delta_\Delta \leq -\Delta) = P\left( \frac{\delta_\Delta}{\sigma_\delta} \leq -\frac{\Delta}{\sigma_\delta} \right) = \Phi\left( -\frac{\Delta}{\sigma_\delta} \right),
    \end{align*}
    where $\Phi(\cdot)$ denotes the cumulative distribution function of the standard normal distribution.  
    To ensure the probability of sign flip is at most 5\%, we require $\Phi\left( -\frac{\Delta}{\sigma_\delta} \right) \leq 0.05$.
    Since $\Phi^{-1}(0.05) = -1.645$, it follows that $-\frac{\Delta}{\sigma_\delta} \leq -1.645\Leftrightarrow\Delta \geq 1.645 \sigma_\delta$.
    Substituting $\sigma_\delta$ gives:
    \begin{align*}
        \Delta \geq 1.645 \sqrt{ \beta_1^2 \sigma_m^2 + (1-\beta_1)^2 \sigma_g^2 }.
    \end{align*}
    Similarly, when both $\Delta$ and $(\Delta+\delta_\Delta)$ are negative, we have $\Delta \leq -1.645 \sqrt{ \beta_1^2 \sigma_m^2 + (1-\beta_1)^2 \sigma_g^2 }$.
    Therefore, in summary, if $|\Delta| \geq 1.645 \sqrt{ \beta_1^2 \sigma_m^2 + (1-\beta_1)^2 \sigma_g^2 }$, then with at least 95\% probability, $\mathrm{sign}(\Delta)$ remains invariant under quantization, i.e., $\mathrm{sign}(\hat{\Delta}) = \mathrm{sign}(\Delta)$.
    \end{proof}

\begin{remark}
We empirically observe the ratio $\frac{|\Delta|}{\sigma_\delta}$ across different iterations and layers. Our results indicate that in at least 97.9\% of cases, the ratio exceeds 1.645, thereby satisfying the condition stated above. Further details are provided in Appendix~\ref{app:lion}.
\end{remark}

\subsection{Hybrid Feature Quantizer: Accurate Updates of Quantized Weights}
\label{method:quantizer}

\begin{figure*}[t]
    \centering
    \includegraphics[width=0.9\linewidth]{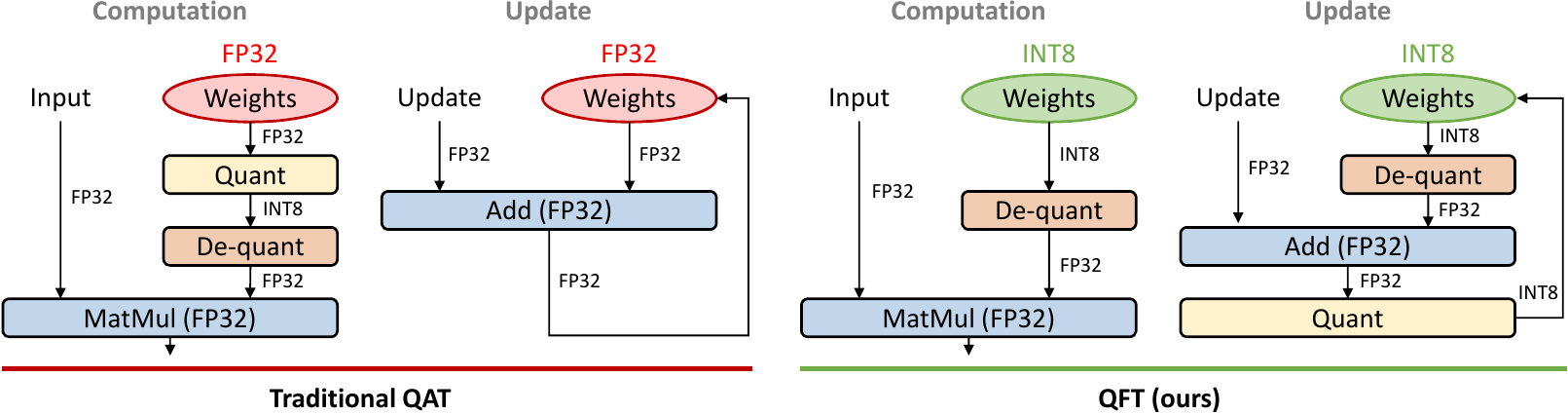}
    \vspace{-0.1cm}
    \caption{Comparison between our QFT and traditional QAT in the computation and update procedures of weights. QAT stores the weights in the floating-point format and adds fake quantization nodes to the computation. Conversely, in our QFT, the weights are stored in the low-precision integer format, which are de-quantized on-the-fly into the floating-point format for computation, resulting in a significant reduction in memory usage.}
\label{fig:qat_vs_qft}
\vspace{-0.3cm}
\end{figure*}

In addition to gradients and momentum, we also store weights in integer format for memory efficiency. This quantized storage is designed purely for training memory optimization and is fundamentally different from traditional QAT \citep{jacob2018quantization}, which inserts fake quantization nodes. In our framework, weights are stored as quantized integers but are de-quantized to floating-point for computation, trading a small amount of extra computation for substantial memory savings. A comparison with QAT is shown in Figure \ref{fig:qat_vs_qft}.

\begin{figure*}[t]
    \centering
    \includegraphics[width=0.9\linewidth]{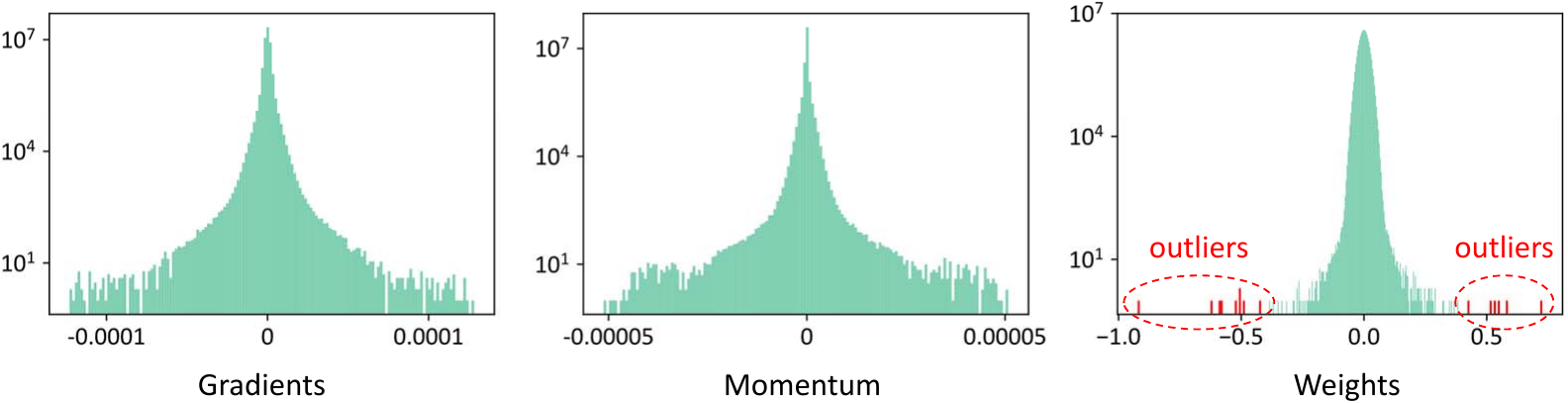}
        \vspace{-0.1cm}
    \caption{Illustration of the model state distributions when training a LLaMA-2-7B model. The weight values are from the final down projection layer, and the gradient and momentum values are fetched on the 200th training step. The gradients and momentum show a canonical centralized distribution with few outliers, while the range of the weights increases by three orders of magnitude and exhibits extreme outliers, posing a significant challenge.}
            \vspace{-0.2cm}
\label{fig:distributions}
\end{figure*}

However, weight quantization is considerably more challenging. By analyzing weight distributions, we make two key observations: (i) a small number of outliers play a crucial role in representation and typically correspond to critical features \citep{kim2023squeezellm,frantar2022gptq,lin2023awq}; (ii) these outliers significantly hinder quantization by greatly expanding the dynamic range of weights, which is about three orders of magnitude larger than that of momentum, as shown in Figure \ref{fig:distributions}. These conflicting properties make outliers particularly difficult to handle.

Fortunately, the sparsity of outliers provides an opportunity. Statistics show that 99\% of the values lie within only 20\% of the full range. This motivates us to adopt a hybrid feature quantizer inspired by SqueezeLLM \citep{kim2023squeezellm}, which preserves the top 1\% sparse but critical features and compacts the remaining dense distribution. Formally, the method is defined as follows:
\begin{align}
    \bm{W} &= \bm{D} + \bm{S}, \\
    \mathrm{s.t.} \enspace \bm{D} = \bm{W} \odot & \mathbb{I}\left( T_{\min} \leq \bm{W} \leq T_{\max} \right),  \notag \\
    \bm{S} = \bm{W} \odot & \mathbb{I}\left( \bm{W} < T_{\min} \ \mathrm{or} \ \bm{W} > T_{\max} \right), \notag
\end{align}
where $\bm{D}$ is a dense matrix containing the central values and $\bm{S}$ is a sparse matrix capturing the outliers. $T_\mathrm{min}$ and $T_\mathrm{max}$ denote outlier thresholds, and $\mathbb{I}(\cdot)$ is the indicator function. This decomposition is numerically simple and highly efficient, introducing negligible training overhead.

The dense matrix is then quantized using the uniform quantizer in Equation~\ref{eq:quant}, while the sparse matrix keeps its values in floating point. Since the outliers account for only a small fraction (1\%), the sparse matrix can be stored in memory-efficient formats such as CSR, incurring negligible memory overhead while preserving critical information.

Moreover, QFT has a key advantage over mixed-precision training, which requires maintaining an FP32 copy of the weights. In mixed-precision FP16 training, forward and backward passes use FP16 weights and gradients, but updates must be applied to an FP32 master copy due to the limited precision and overflow risks of FP16. In contrast, QFT does not rely on simple numerical truncation: it preserves sparse critical features and uniformly quantizes the dense values into integers. As a result, QFT enables stable training without maintaining an additional FP32 weight copy.

\subsection{The Integer Training Pipeline}
\label{method:pipeline}

In this section, we integrate the above strategies into a memory-efficient fine-tuning framework for LLMs. We describe each training stage in detail, including forward propagation, backward propagation, and parameter updates, with particular emphasis on the stack-based gradient flow scheme with $O(1)$ complexity in the integer setting.

\begin{figure}[h]
\vspace{-0.3cm}
    \centering
    \includegraphics[width=0.8\linewidth]{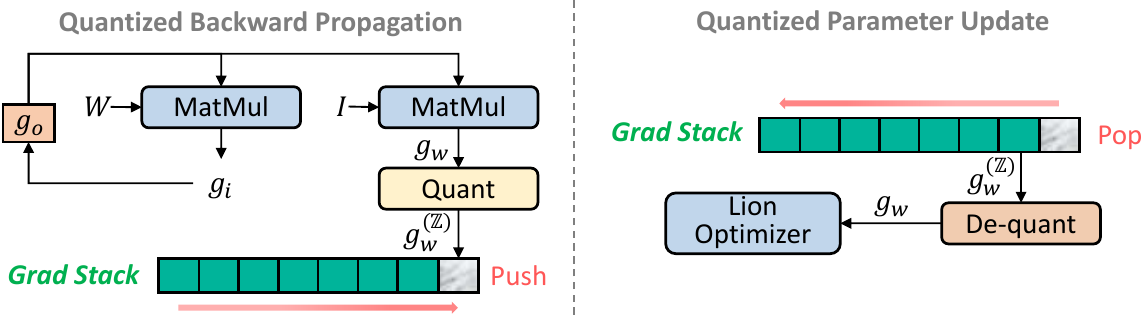}
    \caption{The proposed stack-based gradient flow scheme, which enables storage and $O$(1) complexity access to integer gradients. This effectively eliminates AutoGrad's dependency on floating-point formats, enabling efficient gradient propagation in the context of integer weights.}
\label{fig:grad_flow}
\vspace{-0.3cm}
\end{figure}

\textbf{Quantized Forward Propagation }
In our framework, weights are stored in quantized integer format to reduce memory usage. During forward propagation, these low-precision weights are de-quantized to floating point on the fly, enabling high-precision computation, as illustrated in Figure \ref{fig:qat_vs_qft}.

\begin{minipage}[t]{\textwidth}
  \begin{minipage}[t]{0.51\textwidth}
    \centering
    \vspace{-0.5cm}
    \begin{algorithm}[H]\small
        \caption{Gradient Flow of Quantized Weights}
        \begin{algorithmic}
            \State  {\color{gray}\# $T_l$\;: saved tensors in forward pass of layer $l$}
            \State  {\color{gray}\# $g_o$\;: gradient of the current layer's output}
            \State
            \State $S_g$ $\leftarrow$ \texttt{stack()}
            \For{$l=L,L-1,\cdots, 1$}
                \State $I_l$, $W_l^{(\mathbb{Z})}$ $\leftarrow$ $T_l$
                \State $W_l$ $\leftarrow$ \texttt{{\color[HTML]{CB0000}dequant}($W_l^{(\mathbb{Z})}$)}
                \State \textbf{calculate gradients of $I_l$ and $W_l$}
                \State $g_{i}$ $\leftarrow$ \texttt{matmul($g_o$, $W_l$)}
                \State $g_{w}$ $\leftarrow$ \texttt{matmul($g_o^T$, $I_l$)}
                \State $g_w^{(\mathbb{Z})}$ $\leftarrow$ \texttt{{\color[HTML]{009901}quant}($g_w$)} \hspace*{\fill}{{\color{gray}$\triangleright$ store as INT8}}
                \State \texttt{{\color[HTML]{1E90FF}push}($S_g$, $g_w^{(\mathbb{Z})}$)} \hspace*{\fill}{{\color{gray}$\triangleright$ collect gradient}}
                \State \textbf{assign $g_{o}$ of layer ($l$-1)}
                \State $g_{o}$ $\leftarrow$ $g_{i}$
            \EndFor
        \end{algorithmic}
        \label{alg:grad}
    \end{algorithm}
  \end{minipage}
  \hfill
   \begin{minipage}[t]{0.47\textwidth}
    \centering
    \vspace{-0.5cm}
    \begin{algorithm}[H]\small
        \caption{Quantized Lion Optimizer}
        \begin{algorithmic}
        \State {\color{gray}\# $\beta_1$, $\beta_2$, $\lambda$, $\eta$, $f$\;: optimizer parameters}
        \State {\color{gray}\# $m_l$\;: optimizer momentum of layer $l$}
        \State
        \For{$l=1,2,\cdots, L$}
            \State  $g_w^{(\mathbb{Z})}$ $\leftarrow $ \texttt{{\color[HTML]{1E90FF}pop}($S_g$)} \hspace*{\fill}{{\color{gray}$\triangleright$ retrieve gradient}}
            \State $g_w$ $\leftarrow$ \texttt{{\color[HTML]{CB0000}dequant}($g_w^{(\mathbb{Z})}$)}
            \State $m_l$ $\leftarrow$ \texttt{{\color[HTML]{CB0000}dequant}($m_l^{(\mathbb{Z})}$)}
            \State $W_l$ $\leftarrow$ \texttt{{\color[HTML]{CB0000}dequant}($W_l^{(\mathbb{Z})}$)}
            \State \textbf{update model parameters}
            \State $\Delta \leftarrow \beta_1 m_l + (1-\beta_1)g_w$
            \State $W_l \leftarrow W_l - \eta(\texttt{sign}(\Delta) + \lambda W_l)$
            \State \textbf{update EMA of $g_w$}
            \State $m_l \leftarrow \beta_2 m_l + (1 - \beta_2)g_w$
            \State $m_l^{(\mathbb{Z})}$ $\leftarrow$ \texttt{{\color[HTML]{009901}quant}($m_l$)} \hspace*{\fill}{{\color{gray}$\triangleright$ store as INT8}}
            \State $W_l^{(\mathbb{Z})}$ $\leftarrow$ \texttt{{\color[HTML]{009901}quant}($W_l$)} \hspace*{\fill}{{\color{gray}$\triangleright$ store as INT8}}
        \EndFor
        \end{algorithmic}
        \label{alg:optim}
    \end{algorithm}
  \end{minipage}
\end{minipage}

\textbf{Quantized Backward Propagation }
During backward propagation, the loss is propagated layer by layer and gradients are computed for each parameter. These gradients must be stored in memory to guide subsequent updates. However, in mainstream frameworks such as PyTorch, only floating-point parameters can hold gradients. As a result, gradients for integer-stored weights cannot be computed or stored directly using the standard automatic differentiation (AutoGrad) mechanism.

To address this issue, we design a new gradient flow for integer-stored weights (Figure \ref{fig:grad_flow}, Algorithm \ref{alg:grad}). As in the forward pass, weights are first de-quantized to floating point, and gradients w.r.t. inputs and weights are computed via the chain rule. To handle gradient storage, we further propose a global stack-based retention scheme, where each layer’s weight gradient is pushed onto the stack during backpropagation in reverse forward order.

\textbf{Quantized Parameter Update }
Finally, parameters are updated using the standard Lion optimizer, except that both gradients and momentum are stored in integer format (Algorithm~\ref{alg:optim}). Gradients are retrieved by popping from the global stack in $O(1)$ time. This efficiency stems from our gradient flow design: gradients are pushed onto the stack during backpropagation from the last layer to the first, and popped during optimization in the reverse order, so the current layer’s gradient is always on top of the stack, fully exploiting the LIFO property.

\section{Experiments}

\subsection{Experimental Setup}

\textbf{Baseline Methods }
We evaluate QFT in terms of both training memory and performance.
For training memory, QFT is compared to standard FP32 Adam~\citep{kingma2015adam}, mixed-precision FP16 Adam~\citep{micikevicius2017mixed}, BitsandBytes~\citep{dettmers20218}, and standard FP32 Lion~\citep{chen2302symbolic}.
For training performance, we consider four baselines: \textbf{LoRA}, which is a PEFT method~\citep{hu2021lora}; \textbf{FT-Adam}, which performs full-parameter fine-tuning using FP32 Adam~\citep{kingma2015adam}; \textbf{FT-Lion}, which performs fine-tuning using FP32 Lion~\citep{chen2302symbolic}; \textbf{FT-Bnb}, which uses Bitsandbytes~\citep{dettmers20218} with quantized optimizer states.

Models, benchmarks, datasets, and training details are presented in Appendix \ref{app:exp_setup}.

\subsection{Memory Profile}
The memory usage of different methods are reported in Table \ref{tab:memory}.
All experiments employ gradient checkpointing by default to reduce activation memory.
Under standard Adam~, each model state consumes 25.1GB of RAM. This issue persists in mixed-precision settings.
Lion reduces memory usage by 25\% by tracking only momentum. BitsandBytes further compresses optimizer states via quantization, saving 37GB. 
Our QFT applies full quantization to all model states, requiring only 21.5GB of GPU memory, which is just 21\% of that used by standard Adam.

\begin{table*}[t]\small
\centering
\setlength{\tabcolsep}{4.5pt}
\caption{Memory usage (in GB) when fine-tuning the LLaMA-2-7B model using different methods. We report the  memory profiles, total allocated memory, and peak allocated memory.}
\begin{tabular}{@{}ccccccccc@{}}
\toprule
\multirow{2.5}{*}{Method} &
  \multirow{2.5}{*}{Weights} &
  \multirow{2.5}{*}{Grads} &
  \multicolumn{3}{c}{Optimizer States} &
  \multirow{2.5}{*}{Activation} &
  \multirow{2.5}{*}{Total} &
  \multirow{2.5}{*}{Peak} \\ \cmidrule(lr){4-6}
             &      &      & Weight Copies & Momentum & Variances &      &      &      \\ \midrule
Adam-FP32         & 25.1 & 25.1 & -            & 25.1     & 25.1      & 3.75 & 104  & 129  \\
Adam-FP16 mixed   & 12.6 & 12.6 & 25.1         & 25.1     & 25.1      & 3.75 & 104  & 123  \\
Bitsandbytes & 25.1 & 25.1 & -         & 6.31     & 6.31      & 3.75 & 66.6 & 86.6 \\
Lion-FP32         & 25.1 & 25.1 & -            & 25.1     & -         & 3.75 & 79.1 & 101  \\
\hc QFT (ours)      & \hc 7.42 & \hc 7.06 & \hc -            & \hc 7.06     &  \hc -         & \hc 3.75 & \hc 25.3 & \hc 28.9 \\ \bottomrule
\end{tabular}
\label{tab:memory}
\vspace{0cm}
\end{table*}

\begin{table*}[t]\small
\centering
\setlength{\tabcolsep}{5pt}
\caption{Few-shot performance of different models on various standard benchmarks. Here, the number of shots is aligned to Open LLM Leaderboard~\citep{huggingface_open_llm}. }
\begin{tabular}{@{}cccccccc@{}}
\toprule
Pre-trained & Tuning & Full-param &
  \begin{tabular}[c]{@{}c@{}}ARC-c\\ (25-shot)\end{tabular} &
  \begin{tabular}[c]{@{}c@{}}HellaSwag\\ (10-shot)\end{tabular} &
  \begin{tabular}[c]{@{}c@{}}MMLU\\ (5-shot)\end{tabular} &
  \begin{tabular}[c]{@{}c@{}}TruthfulQA-mc\\ (0-shot)\end{tabular} &
  Average \\ \midrule
\multirow{6}{*}{LLaMA-2-7B} 
& -   & - & 53.1 & 78.6 & 46.9 & 38.8 & 54.4 \\ \cmidrule{2-8} 
& LoRA  &  $\times$ & 53.0 & 78.0 & 47.8 & 45.8 & 56.2 \\
& FT-Adam   & $\checkmark$ & 53.6 & 77.3 & 49.4 & 51.5 & 58.0 \\
& FT-Lion & $\checkmark$ & 53.5 & 77.6 & 49.2 & 51.3 & 57.9 \\
& FT-Bnb & $\checkmark$ & 53.1 & 76.9 & 49.0 & 51.1 & 57.5 \\
& \hc QFT (ours) & \hc $\checkmark$ &  \hc 52.9 &   \hc 76.7   & \hc 48.8 & \hc 51.1 &  \hc 57.4    \\ \midrule \midrule
\multirow{6}{*}{LLaMA-2-13B} 
& -  &  - & 59.4 & 82.1 & 55.8 & 37.4 & 58.7 \\ \cmidrule{2-8} 
& LoRA   & $\times$ & 57.3 & 81.2 & 55.6 & 44.7 & 59.7 \\
& FT-Adam  & $\checkmark$  & 57.0 & 81.2 & 55.8 & 50.9 &  61.2 \\
& FT-Lion & $\checkmark$ & 56.6 & 81.2 & 55.6 & 50.7 & 61.0 \\
& FT-Bnb & $\checkmark$ & 56.0 & 80.9 & 55.4 & 49.4 & 60.4 \\
& \hc QFT (ours) & \hc $\checkmark$ & \hc 56.2  & \hc 81.0 & \hc 55.9 & \hc 48.6 & \hc 60.4   \\ \bottomrule
\end{tabular}
\label{tab:few-shot}
\end{table*}

\subsection{Performance Evaluation}

\textbf{Few-Shot Evaluation }
We perform few-shot performance evaluation across a range of established benchmarks, with results presented in Table \ref{tab:few-shot}. 
On LLaMA-2-7B, QFT boosts the average performance score from 54.4 to 57.4, which comes within 0.6 of full-precision fine-tuning with Adam (FT-Adam), while outperforming LoRA.
We also provide a qualitative analysis in Appendix \ref{app:qualitative}.

\begin{wraptable}{R}{0.55\textwidth}\small
\vspace{-0.4cm}
\setlength{\tabcolsep}{3pt}
\centering
\caption{MT-Bench scores using GPT-4, which can reflect model's conversational abilities.}
\begin{tabular}{@{}cccc@{}}
\toprule
Pre-trained & Tuning  & Full-param    & MT-Bench Score \\ \midrule
\multirow{5}{*}{\makecell{LLaMA-2-7B \\ (Score: 3.83)}}
& LoRA  &  $\times$ & 5.11 \\
& FT-Adam      &   $\checkmark$  & 6.08                   \\
& FT-Lion      &   $\checkmark$  &    6.11           \\
& FT-Bnb      &   $\checkmark$  &     5.94          \\
& \hc QFT (ours)  & \hc $\checkmark$   &   \hc  5.95            \\ \midrule \midrule
\multirow{5}{*}{\makecell{LLaMA-2-13B \\ (Score: 4.69)}}
& LoRA  &  $\times$ & 5.74 \\
& FT-Adam     &  $\checkmark$  & 6.46                   \\
& FT-Lion     &  $\checkmark$  &     6.44             \\
& FT-Bnb     &  $\checkmark$  &       6.30           \\
& \hc QFT (ours) & \hc $\checkmark$  &   \hc 6.27                \\ \bottomrule
\end{tabular}
\label{tab:mt-bench}
\vspace{-0.7cm}
\end{wraptable}
\textbf{MT-Bench Score }
We further adopt MT-Bench, an advanced benchmark designed to evaluate the conversational capabilities of LLMs. It comprises a series of challenging multi-turn, open-ended questions that closely reflect human conversational preferences, with GPT-4 serving as the automatic evaluator. The results are summarized in Table~\ref{tab:mt-bench}.
For 7B models, LLaMA-2 achieves a modest score of 3.83. While LoRA yields a slight improvement, it still falls short by 0.97 compared to full-parameter fine-tuning. Notably, QFT closes this gap, achieving performance comparable to FT-Adam.

\begin{figure*}[t]
\vspace{-0.5cm}
    \centering
    \begin{subfigure}[]{0.33\textwidth}
        \centering
        \includegraphics[width=\textwidth]{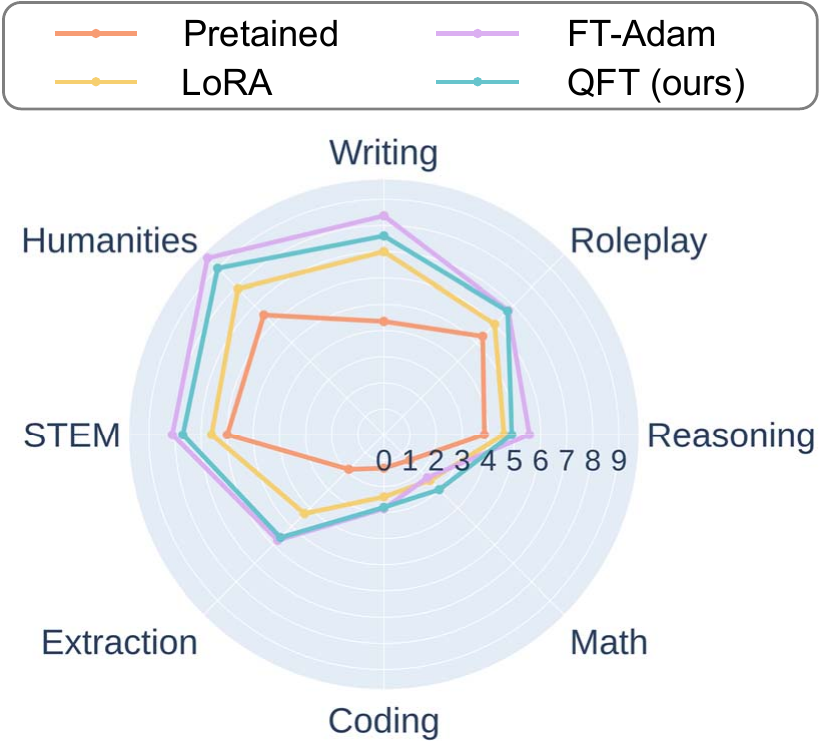}
        \caption{7B models}
    \end{subfigure}
    \qquad \quad
    \begin{subfigure}[]{0.33\textwidth}
        \centering
        \includegraphics[width=\textwidth]{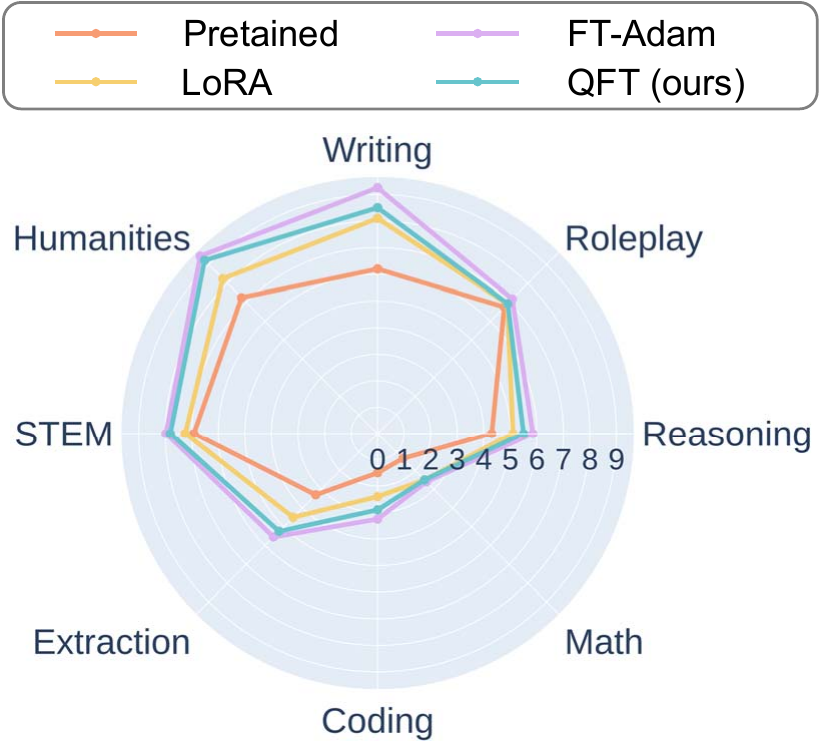}
        \caption{13B models}
    \end{subfigure}
    \caption{Radar charts of each capability in MT-Bench of different tuning method. }
\vspace{-0.7cm}
\label{fig:mt-bench}
\end{figure*}

We also provide radar charts in Figure~\ref{fig:mt-bench}. It demonstrates that QFT delivers comprehensive and consistent improvements across all evaluated metrics compared to the pre-trained LLaMA-2 baseline. 

\begin{wrapfigure}{r}{0.45\textwidth}
\vspace{-0.6cm}
    \centering
    \includegraphics[width=0.42\textwidth]{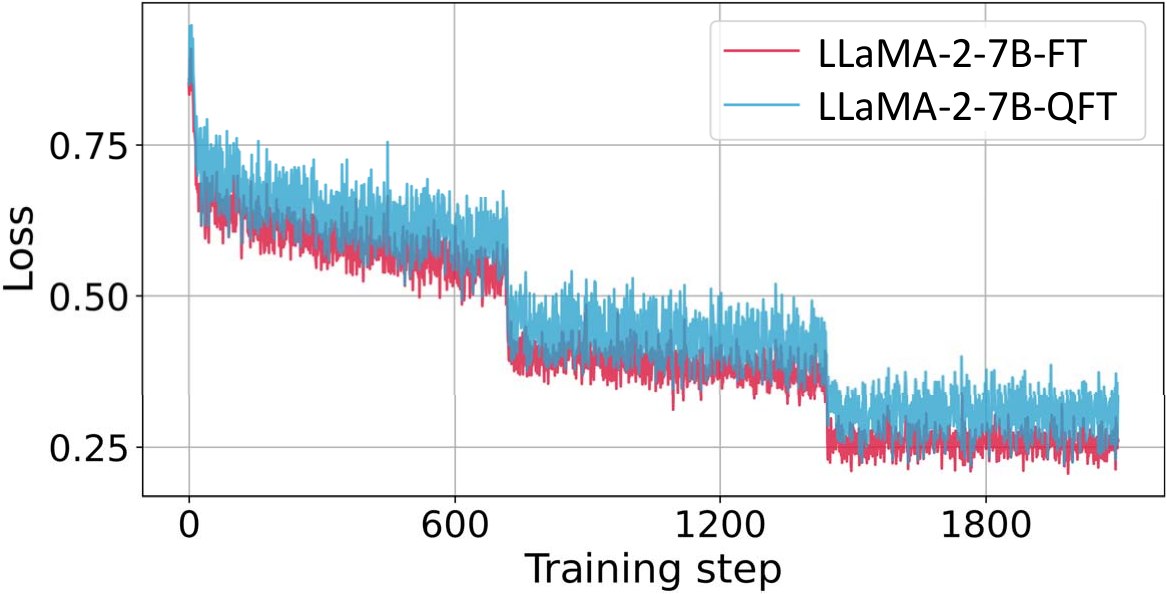}
    \caption{Comparison of training loss curves.}
\label{fig:loss_curve}
\vspace{-0.5cm}
\end{wrapfigure}

\textbf{Training Throughput and Convergence }
We measure the average time per 1000 training steps. Due to quant-dequant overhead, QFT incurs a 1.2-1.3$\times$ increase in training time compared to LLaMA-2-FT with FP32 Adam. This time–memory trade-off is considered significant, especially under constrained memory budgets.
Moreover, Figure \ref{fig:loss_curve} compares the training losses of QFT and LLaMA-2-FT, showing that QFT achieves comparable convergence.

\vspace{-0.5cm}
\section{Conclusion}
\vspace{-0.4cm}
In this paper, we propose QFT, a memory-efficient framework for full-parameter fine-tuning of LLMs under quantized training. QFT quantizes all training states, including weights, gradients, and optimizer states, into INT8, enabling end-to-end training with substantially reduced memory usage while preserving performance. To ensure training stability, we introduce two key techniques: a quantization-robust optimizer (Lion) for momentum and gradients, and a hybrid feature quantizer that preserves sparse critical weight features. We further propose a stack-based gradient flow with $O(1)$ complexity to support efficient integer-domain training.
Experiments show that QFT reduces model state memory to 21\% of standard FP32 fine-tuning, enabling LLaMA-2-7B to be fine-tuned on commodity GPUs such as an A6000 with less than 30GB memory.

\bibliography{iclr2026_conference}
\bibliographystyle{iclr2026_conference}

\clearpage
\appendix

\section{Experimental Setup}
\label{app:exp_setup}

\textbf{Models and Benchmarks }
We conduct evaluation of the proposed QFT by fine-tuning LLaMA-2~\citep{touvron2023llama}, including the 7B and 13B versions.
The few-shot performance of fine-tuned models is evaluated on various standard benchmarks, including ARC~\citep{clark2018think}, HellaSwag~\citep{zellers2019hellaswag}, MMLU~\citep{hendrycks2020measuring}, and TruthfulQA~\citep{lin2021truthfulqa}. 
All results are obtained using the Language Model Evaluation Harness tool~\citep{eval-harness}. In addition, we also use MT-Bench~\citep{zheng2023judging} with GPT-4 scores to evaluate the conversational abilities.

\textbf{Dataset Preparation }
We use a dataset comprising 94.1K ShareGPT entries~\citep{shareGPT-data, shareGPT}, which capture user interactions with ChatGPT. Following FastChat~\citep{vicuna2023}, we convert HTML content to Markdown, remove non-English conversations, and split long dialogues into segments of up to 2048 tokens. For a fair comparison, all baseline methods are replicated using the same dataset as described above.

\textbf{Training Details }
The PEFT method LLaMA-2-LoRA follows the settings: the rank is 8, the alpha is 16, the dropout is 0.05, the learning rate is 2e-5, and the total number of epochs is 3. For full-parameter fine-tuning, all methods follow the same settings: the global batch size is 128, the learning rate is 2e-5, and the total number of epochs is 3.
In QFT, we apply channel-wise quantization for all quantizers of model states. The threshold $T$ in the dense-and-sparse quantizer is obtained from 1\% of the distribution range (see Appendix \ref{app:thresholds} for details).

\section{The Standard Lion Procedure}
\label{app:sec1}
Here, we show the standard Lion procedure with full-precision calculations in Algorithm \ref{app-alg:optim}.

\begin{algorithm}[h]
    \caption{Lion Optimizer}
    \begin{algorithmic}
    \State \textbf{given} $\beta_1, \beta_2, \lambda, \eta, f$
    \State \textbf{initialize} $\theta_0, m_0 \gets 0$
    \For{$t=1,2,\cdots, T$}
        \State $g_t \gets \nabla_\theta f(\theta_{t-1})$
        \State \textbf{update model parameters}
        \State $c_t \gets \beta_1 m_{t-1} + (1 - \beta_1) g_t$
        \State $\theta_t \gets \theta_{t-1} - \eta ( \text{sign}(c_t) + \lambda \theta_{t-1})$
        \State \textbf{update EMA of} $g_t$
        \State $m_t \gets \beta_2 m_{t-1} + (1 - \beta_2) g_t$
    \EndFor
    \State \textbf{return} $\theta_t$
    \end{algorithmic}
    \label{app-alg:optim}
\end{algorithm}

\section{Analysis of Values of $\frac{|\Delta|}{\sigma_\delta}$}
\label{app:lion}

\begin{figure}[h]
    \centering
    \includegraphics[width=0.5\linewidth]{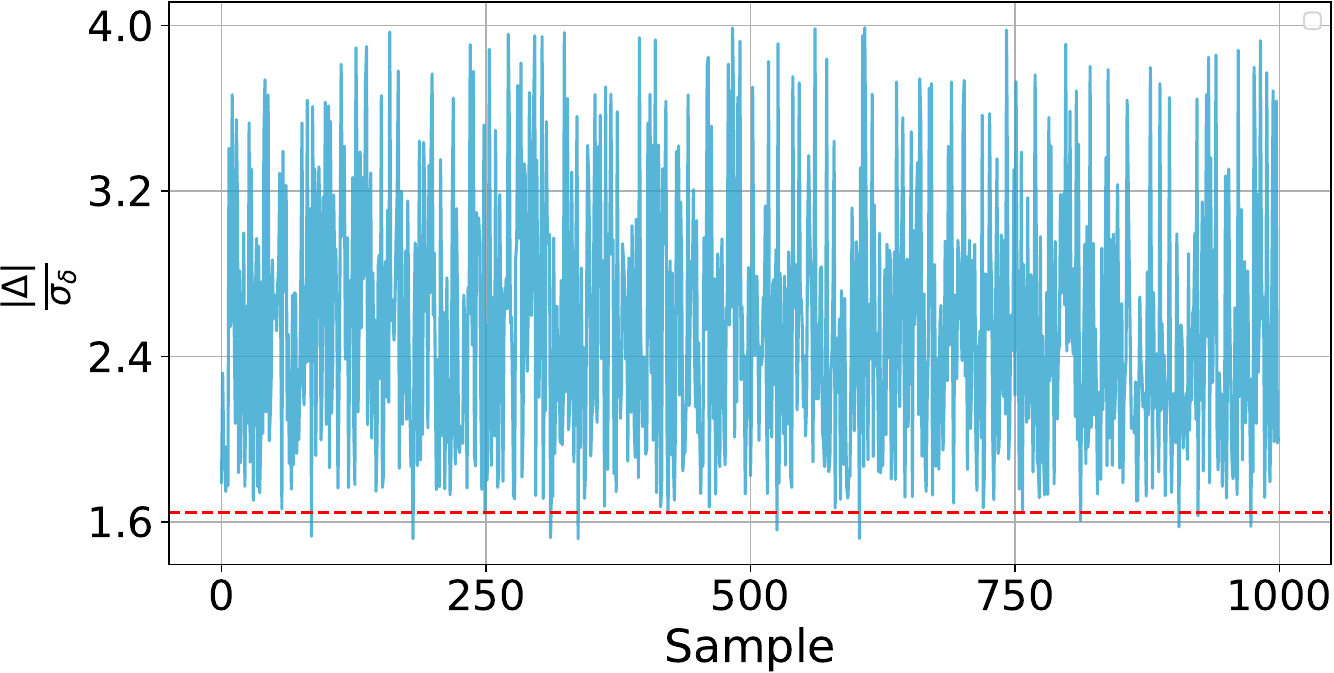}
    \caption{Display of 1000 samples sampled at the 200th training step of the final down projection layer.}
\label{app-fig:lion_proof}
\end{figure}

Lemma 1 shows when $|\Delta|\ge 1.645\sqrt{\beta_1^2\delta_m^2+(1-\beta_1)^2\delta_g^2}$, there is a 95\% probability that sign($\Delta$) is invariant to quantization. Here, we experimentally verify that the condition hold well. Specifically, we sample $\frac{|\Delta|}{\sigma_\delta}$ in different iterations of different layers. $\sigma_\delta$ is obtained by maximum likelihood estimation based on the observed value. The results are presented in Table \ref{app-tab:lion}. As we can see, at least 
97.9\% of cases are greater than 1.645, satisfying the above condition. We also randomly select 1000 samples and visualize their values, as shown in Figure \ref{app-fig:lion_proof}.

\begin{table}[h]
\centering
\setlength{\tabcolsep}{10pt}
\caption{Percentage distribution of $\frac{|\Delta|}{\sigma_\delta}$ in different iterations of different layers. We can see that more than 97.9\% of cases are greater than 1.645.}
\begin{tabular}{@{}cccc@{}}
\toprule
Layer               & Iteration & $\frac{|\Delta|}{\sigma_\delta}<1.645$ & $\frac{|\Delta|}{\sigma_\delta}\ge1.645$ \\ \midrule
\multirow{2.5}{*}{0}  & 200 & 1.6\% & 98.4\% \\ \cmidrule(l){2-4} 
                    & 1200 & 1.3\% & 98.7\% \\ \midrule
\multirow{2.5}{*}{15} & 200 & 1.1\% & 98.8\% \\ \cmidrule(l){2-4} 
                    & 1200 & 1.1\% & 98.9\% \\ \midrule
\multirow{2.5}{*}{31} & 200 & 2.1\% & \cellcolor{teal!20} 97.9\% \\ \cmidrule(l){2-4} 
                    & 1200 & 1.9\% & 98.1\% \\ \bottomrule
\end{tabular}
\label{app-tab:lion}
\end{table}

\section{Discussion on Outlier Thresholds of Weight Quantizer}
\label{app:thresholds}
In this section, we discuss the selection and updating strategies for outlier thresholds in dense-and-sparse quantizers. We first report the memory and accuracy of dense-and-sparse quantizers using different percentage thresholds, and the results are shown in Table \ref{app-tb:thresh}. The accuracy, i.e., the degree of distributional approximation of the quantizers, is evaluated by $L_2$ distance between de-quantized weights $\hat{\bm{W}}$ and full-precision weights $\bm{W}$, where the quantized weights are from the final down projection layer.

\begin{table}[h]
\centering
\caption{Comparison of memory (in GB) and accuracy of dense-and-sparse quantizers using different percentage thresholds for weights. Here, accuracy is measured by $L_2$ distance between de-quantized $\hat{\bm{W}}$ and full-precision $\bm{W}$.}
\setlength{\tabcolsep}{8pt}
\begin{tabular}{@{}cccccc@{}}
\toprule
Percentile  & 0 & 0.45\% & 1.0\% & 3.0\% & 5.0\% \\ \midrule
Memory & 7.06 &  7.23  & \cellcolor{teal!20}7.42 & 8.23 & 9.16  \\
$L_2$ Distance  & 436  & 0.846 &  \cellcolor{teal!20}0.619   & 0.566  & 0.479 \\ \bottomrule
\end{tabular}
\label{app-tb:thresh}
\end{table}

The benefits of employing matrix decomposition in dense-and-sparse quantizers are readily evident from the results. When the percentile is set to 0, the quantizer no longer filters out the outliers and degenerates into a standard uniform quantizer, resulting in intolerable quantization errors.
A noteworthy value suggested in SqueezeLLM~\citep{kim2023squeezellm} is 0.45\%. However, it's essential to acknowledge that this tight bound is primarily designed for inference scenarios. To accommodate potential fluctuations during training, some level of relaxation in the thresholds is necessary. To this end, we assess the relaxed constraints at percentiles of 1.0\%, 3.0\%, and 5.0\%. The findings reveal that, although the 3\% and 5\% percentiles yield a slight boost in accuracy, they also incur higher memory usage. Consequently, we advocate for the 1\% percentile as a more favorable choice for achieving a balance between memory conservation and accuracy.

We also explore the update frequency of the thresholds. In order to maintain a high training throughput, we adopt a strategy of lazy updates, meaning that the thresholds set in the first iteration are reused in subsequent iterations. This approach is viable due to the smooth numerical updates of the weights during the fine-tuning process across neighboring iterations. Additionally, to our surprise, we discover that the initial thresholds remain effective for an entire epoch. As a result, we only update the thresholds once at the beginning of each epoch.

\section{Qualitative Analysis of Conversational Abilities}
\label{app:qualitative}
In the experimental section, we demonstrate the effectiveness of QFT through quantitative evaluations of LLMs' performance on a variety of benchmarks. Despite the large number of benchmarks, the results of the machine evaluations are not considered to be comprehensive, and in some cases, their reliability may be subject to scrutiny, e.g., the "shortcut" issue. Consequently, in the following, we intuitively show and compare examples generated by LLMs, aiming to further verify the conversational abilities through qualitative analysis.

\textbf{\texttt{\color[HTML]{CB0000}[Writing Ability]}}

\textbf{\emph{Question}}: \ul{Write a captivating movie review for a recently released science fiction film, discussing its plot, characters, and special effects.}
 
\textbf{LLaMA-2-7B-FT-Adam}: 
"The Last Voyage" is a thrilling science fiction film that follows a group of astronauts on a mission to explore a distant planet. When they arrive, they discover that the planet is inhabited by a mysterious alien race. As they try to communicate with the aliens and gather data, they realize that their mission may be more dangerous than they ever imagined. With stunning special effects and a gripping plot, "The Last Voyage" is a must-see for science fiction fans.

\textbf{LLaMA-2-7B-QFT}:
"Intergalactic Intrigue" is a thrilling science fiction film that takes viewers on a journey through the stars. The film follows a group of space explorers as they embark on a mission to uncover the secrets of an ancient alien civilization. 
The film's special effects are top-notch, with stunning visuals that transport viewers to a world of wonder and discovery. The characters are well-developed and likable, with each member of the crew bringing their own unique skills and personalities to the table.

\textbf{\texttt{\color[HTML]{CB0000}[Roleplay Ability]}}

\textbf{\emph{Question}}: \ul{As a superhero, how would you explain your origin story to a curious child?}
 
\textbf{LLaMA-2-7B-FT-Adam}: I was just like any other kid, until one day I was exposed to a strange chemical that gave me incredible powers. I learned to control the elements and use them to protect the innocent. I became known as the Elemental Guardian, and I've been fighting crime and saving the day ever since.

\textbf{LLaMA-2-7B-QFT}: I was just like any other kid, playing in the park and having fun. But one day, I was hit by a strange ray that gave me superpowers. I could fly, turn invisible, and even control the weather! I decided to use my powers to help people and fight crime. That's how I became the superhero you know and love!

\textbf{\texttt{\color[HTML]{CB0000}[Common-sense Ability]}}

\textbf{\emph{Question}}: \ul{In a world where automation is becoming increasingly prevalent, is it more important to prioritize job creation or technological progress?}

\textbf{LLaMA-2-7B-FT-Adam}: Job creation is essential for a thriving economy, as it provides individuals with employment opportunities and a source of income. While technological progress is important for driving innovation and improving efficiency, it is crucial to balance technological advancements with the creation of new jobs. This can be achieved by investing in education and training programs that help workers develop the skills necessary to adapt to new technologies and by promoting policies that support small businesses and entrepreneurship. By prioritizing both job creation and technological progress, we can ensure a more stable and prosperous future for all.

\textbf{LLaMA-2-7B-QFT}: In the short term, job creation is a critical concern for policymakers and businesses. The unemployment rate remains high, and many people are struggling to find work. However, in the long term, technological progress is essential for driving economic growth and improving the standard of living for everyone.
Automation has the potential to increase productivity and efficiency, which can lead to higher wages and better working conditions for employees. However, it can also lead to job displacement, particularly in industries where automation is most prevalent.

\textbf{\texttt{\color[HTML]{CB0000}[Fermi Ability]}}

\textbf{\emph{Question}}: \ul{How many snowflakes fall during a typical winter? Try to explain your answer. Your explanation should take the reader through your reasoning step-by-step.}

\textbf{LLaMA-2-7B-FT-Adam}: It is difficult to determine the average number of snowflakes that fall during a typical winter, as it can vary greatly depending on location, climate, and other factors. However, it is estimated that on average, a single snowflake can have up to 100,000 ice crystals per flake. Over the course of a typical winter, this could result in billions of snowflakes falling from the sky.

\textbf{LLaMA-2-7B-QFT}: To answer this question, we need to consider the average snowfall for a typical winter in a given location.
In the United States, for example, snowfall tends to be heaviest in the northern states, where the average snowfall for a typical winter is around 60 inches. In the southern states, snowfall is much less common and the average snowfall for a typical winter is around 10 inches.

\textbf{\texttt{\color[HTML]{CB0000}[Counterfactual Ability]}}

\textbf{\emph{Question}}: \ul{What if the Internet had been invented during the Renaissance period?}

\textbf{LLaMA-2-7B-FT-Adam}: The Renaissance would have been a much different time period if the internet had been invented. The flow of information would have been much faster and more efficient, leading to a greater exchange of ideas and innovations. The world would have been connected in ways never before seen, allowing for the rapid spread of new ideas and technologies.

\textbf{LLaMA-2-7B-QFT}: 
The Renaissance was characterized by a renewed interest in classical learning and the arts, and the Internet would have provided a way for people to access and share this knowledge. It would have also facilitated the exchange of ideas and the development of new technologies, leading to even greater advancements in the fields of science, medicine, and engineering.

\textbf{\texttt{\color[HTML]{CB0000}[Generic Ability]}}

\textbf{\emph{Question}}: \ul{Can you explain the basics of quantum computing?}

\textbf{LLaMA-2-7B-FT-Adam}: Quantum computing is a type of computing that uses quantum bits (qubits) to perform calculations, unlike traditional computers that use binary.

\textbf{LLaMA-2-7B-QFT}: Quantum computing is a type of computing that uses quantum-mechanical phenomena, such as superposition and entanglement, to perform operations on data. These phenomena allow quantum computers to perform certain calculations much faster than classical computers.

\textbf{\texttt{\color[HTML]{CB0000}[Knowledge Ability]}}

\textbf{\emph{Question}}: \ul{Explain the process of natural selection and how it contributes to the evolution and adaptation of species.}

\textbf{LLaMA-2-7B-FT-Adam}: Natural selection is the process by which species adapt to their environment through the survival and reproduction of individuals with favorable traits. Over time, these favorable traits become more common in the population, leading to the evolution of new species.

\textbf{LLaMA-2-7B-QFT}: Natural selection is a process where species that are better adapted to their environment tend to survive and reproduce more than those that are less adapted. This leads to the survival of the fittest and the gradual evolution of species over time.

\end{document}